\documentclass[11pt,a4paper,twocolumn]{article}

\usepackage[T1]{fontenc}
\usepackage[utf8]{inputenc}
\usepackage{microtype}
\usepackage{times}
\usepackage[a4paper,top=1in,bottom=1in,left=0.95in,right=0.95in,columnsep=0.31in]{geometry}
\usepackage{graphicx}
\usepackage{booktabs}
\usepackage{amsmath,amssymb}
\usepackage{hyperref}
\usepackage{url}
\usepackage{xcolor}
\usepackage{caption}
\usepackage{subcaption}
\usepackage{multirow}
\usepackage{array}
\usepackage{float}
\usepackage{textcomp}
\usepackage{seqsplit}
\usepackage{natbib}
\usepackage{titlesec}

\hypersetup{
    colorlinks=true,
    linkcolor=blue!60!black,
    citecolor=blue!60!black,
    urlcolor=blue!60!black,
    breaklinks=true
}

\newcommand{\eps}{\varepsilon}
\newcommand{\xtimes}{\times}

\titlespacing*{\section}      {0pt}{2.0ex plus 0.5ex minus 0.2ex}{1.0ex plus 0.2ex}
\titlespacing*{\subsection}   {0pt}{1.5ex plus 0.4ex minus 0.2ex}{0.5ex plus 0.2ex}
\titlespacing*{\paragraph}    {0pt}{0.6ex plus 0.2ex minus 0.1ex}{0.5em}
\title{What, Where, and How: Disentangling the Roles of Task, Language, and Model in Code Model Representations}

\author{
    Piotr Wilam \\
    University College London \\
    \texttt{piotrwilam@gmail.com}
}

\date{}

\begin{document}

\twocolumn[
\begin{@twocolumnfalse}
\maketitle
\begin{abstract}
\noindent
When two language models are trained independently, on different data, do they come to represent the same thing in the same way? We answer this question for code. We extend a recently introduced concept-circuit extraction method to a 2$\xtimes$2 design --- two languages (Python, Rust) crossed with two models (Qwen2.5-Coder-7B, DeepSeek-Coder-V1-6.7B) --- and measure a complete inventory of grammatical concepts (58 Python, 57 Rust) identically in all four cells. This is the smallest design that can separate what depends on the task, what depends on the language, and what depends on the model.

\vspace{0.5em}\noindent
The answer splits into three parts. \emph{What} earns dedicated circuitry is set by the task: the two models agree on which concepts receive circuits (Spearman $\rho = 0.638$ for Python, $0.673$ for Rust, both $p < 10^{-7}$). \emph{Where} those circuits sit is set by the model: Qwen processes concepts in a late band ($\sim$L17--19), DeepSeek at L6--7 --- a 12--13 layer gap that holds for both languages. \emph{How} circuits grow across layers is also set by the model: Qwen gives its atomic concepts an early spike that DeepSeek does not. The question ``are circuits universal?'' therefore has no single answer. It is yes for What and no for Where and How: universality is a property of representational content, not of computational organisation.

\vspace{0.5em}\noindent
None of this quantitative structure was fixed in advance. The cross-model agreement could have landed anywhere between independence and identity; it lands at $\rho \approx 0.65$. Rust constructs receive 2--3$\xtimes$ more concept-specific circuitry than their Python equivalents, in both models. Both models share neurons between the two languages (6 of 7 and 7 of 7 paired constructs), and DeepSeek shares 1.94$\xtimes$ more than Qwen --- a direction no prior result predicts. And Qwen binds nine keywords of Rust's type-and-trait machinery into a single tight neuron cluster (within-group Jaccard $0.535$ against a permutation null of $0.112$, $p < 0.001$), recovering a semantic dimension that is invisible in surface syntax. Causal ablation and linear probes confirm that the extracted circuits are functional.

\vspace{0.5em}\noindent
All claims are scoped to this 2$\xtimes$2. Two models are enough to show that the axes separate; whether the resulting per-model profile predicts a third model is the designed next test.
\end{abstract}
\vspace{1.5em}
\end{@twocolumnfalse}
]

\section{Introduction}
\label{sec:intro}

When a model encounters a \texttt{for} loop in Python and a \texttt{for} loop in Rust, does it use the same neurons? When two independently trained models both process Python's \texttt{import}, do they allocate circuitry of comparable depth, in the same layers, with the same growth pattern? Three questions about abstraction are fundamental to interpretability: how an abstraction is formed, where it lives, and how it is processed. This paper addresses the second and third; formation is the subject of separate developmental work.

The obstacle to answering them has been methodological, not conceptual. Circuit discovery traces one behaviour in one model \citep{Wang2023,Conmy2023}. Probing tests one hypothesis at a time \citep{Tenney2019,Belinkov2022}. Sparse autoencoders discover features bottom-up, but in a basis learned per model, so two models' features have no built-in correspondence \citep{Bricken2023,Cunningham2023}. What the question needs is different: a single measurement, defined identically for any model and any language, taken over a complete inventory of constructs. This paper builds that measurement and uses it.

We extend the extraction-and-marginalisation method of Wilam (2026), developed on a sparse 8-layer Python transformer, to a 2$\xtimes$2 design across Python and Rust on two dense production models. The unit of analysis is a structured inventory: 58 testable Python concepts and 57 testable Rust concepts, measured comparably in all four cells. Formal languages are the right setting to develop the method, because the grammar fixes the inventory of constructs in advance; natural language, where the inventory is messier but the questions matter most, is the eventual target.

\paragraph{Why the outcome was not a foregone conclusion.} A reasonable prior says that models trained on similar data should agree on content while differing in mechanism, so the \emph{direction} of our headline result could have been guessed. But a direction is not a result, and before this study the expectation had no evidential basis: no prior work had measured a structured inventory of constructs comparably across two independently trained models. Nobody knew whether cross-model agreement on ``what earns circuitry'' would land near $\rho = 0.3$, $0.65$, or $0.9$ --- three outcomes with very different implications. The data picks the middle one: a conserved ranking, far from identity.

Nothing in prior work fixes the direction of the model differences either. That DeepSeek shares 1.94$\xtimes$ more neurons across languages than Qwen; that the two models' processing bands sit 12--13 layers apart, in a consistent order; that Rust earns 2--3$\xtimes$ more concept-specific circuitry than Python in both models --- each could have gone the other way, or vanished. And the strongest structural finding is one nobody predicted: six independent measurements of the two models do not scatter into six separate differences. They collapse into two coherent processing styles (\S\ref{sec:discussion}). That is a statement about the \emph{dimensionality} of cross-model variation, not the truism that different models differ.

\paragraph{What the decomposition enables.} The axes have direct practical content. Because What transfers, a concept inventory measured on one model is a usable prior for another: concept-level results are portable. Because Where and How do not transfer, any layer-indexed technique --- probing at a fixed depth, patching at a peak layer, editing a located circuit --- is model-specific and must be re-localised per model; our per-model band measurements say where. A practitioner porting an interpretability result across models should ask which axis it lives on.

The central result is that the two models behave as carriers of two distinct processing styles: they agree on which constructs earn dedicated circuitry but diverge on where and how those circuits are organised. The What/Where/How dissociation organises the evidence --- and one discovery shows the method reaching past the syntax it was built to measure: Qwen binds Rust's type-and-trait machinery into a single neuron cluster, a semantic dimension absent from the surface code.

\paragraph{Roadmap.} \S\ref{sec:background} situates the work and summarises the extraction method self-containedly. \S\ref{sec:method} walks through the pipeline, defining each term as it arises. \S\ref{sec:what_where_how} establishes the dissociation: What is conserved, Where diverges, How diverges. \S\ref{sec:lang} develops the language axis. \S\ref{sec:arch} develops organisation that survives the architecture change, including the type-trait cluster. \S\ref{sec:validation} validates the decomposition causally and geometrically. \S\ref{sec:discussion} synthesises the measurements into a two-styles fingerprint and states the predictions it licenses.

\section{Background and Related Work}
\label{sec:background}

\paragraph{Circuit discovery.} Mechanistic interpretability reverse-engineers networks into human-understandable components --- from curated case studies such as the indirect-object-identification circuit \citep{Wang2023} to automated discovery via activation patching \citep{Conmy2023} and its refinements \citep{HeimersheimNanda2024}. These methods start from a behaviour and find its circuit: one behaviour, one model at a time. Wilam (2026) inverts the stance --- start from a structured concept inventory and find the circuitry for each concept. The inversion is what makes cross-model comparison possible, because the inventory, not the behaviour, becomes the common yardstick.

\paragraph{Features and superposition.} Individual neurons respond to multiple unrelated features \citep{Elhage2022}. Sparse autoencoders address this by learning overcomplete feature bases in which representations are more interpretable \citep{Bricken2023,Cunningham2023,Templeton2024}, and neuron-level studies locate interpretable units directly \citep{Gurnee2023}. But SAE features are defined per model: two models' dictionaries carry no a-priori correspondence, which is exactly what a cross-model comparison cannot work with. Our binary neuron-mask measurement is coarser, and deliberately so --- it is defined identically for any model with an MLP, which is what the 2$\xtimes$2 requires. \S\ref{sec:validation:probes} quantifies what the coarseness costs.

\paragraph{Universality.} Whether independently trained networks converge on the same internal structures is a long-standing question \citep{Olah2020,Chughtai2023}. Recent work pursues it at scale: universal neurons recur across GPT-2 training runs \citep{Gurnee2024}; the Platonic Representation Hypothesis argues that models converge toward a shared representation of reality as they grow \citep{Huh2024}; and feature-space and mechanistic similarity across architectures are active measurement targets \citep{Lan2024,Wang2024}. This literature mostly poses universality as a single yes/no question; our contribution is to split it into axes that receive different answers.

\paragraph{Cross-model comparison.} Cross-model representational comparison is commonly done with similarity measures such as CKA \citep{Kornblith2019}. We instead compare a structured concept inventory directly. The trade is deliberate: an inventory comparison is blunter than a geometry comparison, but it separates agreement on concept \emph{identity} from divergence in layer \emph{placement} --- the distinction the universality question turns on, and one a single similarity score cannot express.

\paragraph{Probing, and code models.} Linear probes detect syntactic structure \citep{Tenney2019,Belinkov2022}; they ask whether information is decodable, a different question from which neurons carry it. \S\ref{sec:validation:probes} uses both views and measures where they agree. For code specifically, prior work probes syntax trees \citep{Wan2022}, studies relation decoding \citep{Hernandez2024}, and localises code syntax to early layers in a single model \citep{Yin2025}. Our cross-model design tests that last kind of claim directly: we find the band is model-determined --- early in DeepSeek, late in Qwen --- so ``where code structure lives'' is a per-model fact, not a general one.

\paragraph{Cross-lingual sharing.} Multilingual models share representations across natural languages \citep{Pires2019,Conneau2020,Muller2021}, and recent mechanistic work sharpens the picture: models reuse the same circuits for shared syntactic processes while allocating language-specific components for language-unique ones \citep{Zhang2025}, with language-specific neurons concentrated in the top and bottom layers \citep{Tang2024}, and shared concept spaces emerging over multilingual training \citep{Korner2026}. We extend this line from natural to \emph{formal} languages --- and add the cross-model axis it lacks, asking not only whether circuits are shared across languages but whether the \emph{amount} of sharing is a language fact or a model fact (\S\ref{sec:lang:sharing}: the ranking is shared; the amount is not).

\paragraph{The concept-circuit method, self-contained.} Wilam (2026) introduced the method on a sparse 8-layer Python model; since this paper stands on it, we summarise it fully. For each concept in a language's inventory --- say the Python AST node \texttt{For} --- one writes many prompts that all contain the concept while varying everything else. A neuron that fires on all of them is responding to the concept, not to any incidental feature. The intersection of the active-neuron sets over the prompt family is therefore a marginalisation: it integrates out the varied dimensions and leaves the concept. A second family --- checker prompts --- places the concept's keyword token outside its structural role, and the same construction yields the \emph{token's} circuit. Comparing the two separates the abstraction from the surface form. On the sparse model, the original paper established three things: the intersections stay populated (random sets of the same density shrink to empty); they recover known conceptual structure; and an \emph{atomicity} distinction --- modular keywords with no nested body versus body-scoped constructs --- organises the circuits' temporal behaviour. Everything downstream in this paper is this method, held fixed, applied to four (model, language) cells.

\section{Methodology}
\label{sec:method}

The pipeline has five stages.

\paragraph{Prompt generation.} For each concept, two prompt sets: \emph{concept prompts} place the construct in its structural role; \emph{checker prompts} place the keyword token outside that role (in strings, comments, variable names). Fifty prompts per (concept, context) pair, with variance injected along lexical, structural, and padding dimensions, so marginalisation washes out incidental features. Formal languages give a finite a-priori inventory; we take the testable subset --- constructs whose keyword token can also appear outside its structural role, which the contrast requires --- excluding constructs whose token is inseparable from the concept (Rust ownership operators) and tokenless constructs. This yields 58 testable Python and 57 testable Rust concepts; for the cross-language comparison (\S\ref{sec:lang:sharing}) we identify seven construct pairings among them.

\paragraph{Extraction.} Each prompt passes through the model once; forward hooks record the MLP output at the last token position of every layer.

\paragraph{Binarisation.} A neuron is active if its absolute activation exceeds $\eps$. We sweep $\eps \in \{0.001, 0.1, 0.5\}$ to probe permissive to strict: at $0.001$ circuits are trivially full, at $0.1$ concept-only fractions are near-zero, and structural signal emerges at $\eps = 0.5$, which we use throughout. A second filter would retain neurons active in $\geq 80\%$ of prompt variations, but for these dense SwiGLU models every neuron passing $\eps$ fires on all variations, so it is inert and the sweep reduces to one parameter. (Artifact names and figure panels label this filter \texttt{cons=0.8}, its fixed 80\% setting; it does no work here.) The all-or-nothing firing means the selected neurons are stable concept detectors, robust to the injected variance, rather than responders to incidental prompt features.

\paragraph{Marginalisation.} For the Python AST concept \texttt{For}, prompts place a \texttt{for} loop over many builtins (over a \texttt{list}, a \texttt{dict}, a \texttt{range}); the universal circuit for \texttt{For} is the intersection of their active-neuron sets --- neurons that fire regardless of which builtin appears --- so intersecting marginalises the builtin out and leaves the concept. Builtin concepts are marginalised analogously across the AST constructs they appear inside. (``AST node'' and ``builtin'' are Python's categories; Rust constructs are marginalised across their complementary set.) Were a concept not encoded as a stable unit, the intersection would shrink toward empty, as for random sets of the same density; that it stays populated is what makes a universal circuit meaningful.

\paragraph{Decomposition.} The universal mask $A$ is compared against the checker mask $B$: concept-only $= A \setminus B$, shared $= A \cap B$, token-only $= B \setminus A$. The \emph{concept fraction} $|A \setminus B| / |A|$ is the central per-concept measurement. The result is a dataset indexed by concept, layer, model, and language.

The analyses in \S\ref{sec:what_where_how}--\S\ref{sec:validation} use standard tools applied to this dataset: Spearman correlation for cross-model agreement (\S\ref{sec:what}); Ward-linkage clustering with a permutation null over pairwise Jaccard similarity (\S\ref{sec:clustering}); zero-ablation against a size-matched null (\S\ref{sec:validation:ablation}); and logistic-regression probes with cosine comparison (\S\ref{sec:validation:probes}). The pipeline separates into three layers --- extraction (run once, frozen artifacts), analysis (seconds, on artifacts), and presentation (one script per figure) --- all released; because extraction is frozen, every number and figure regenerates from the analysis layer without rerunning the model (Appendix~R).

\section{The What/Where/How Dissociation}
\label{sec:what_where_how}

\subsection{Calibration}
\label{sec:calibration}

The method recovers known positives and negatives. The atomicity super-cluster from Wilam (2026) replicates: in Qwen the Python \texttt{two\_phase} concepts are the six-concept set \{Assert, Break, Continue, Import, ImportFrom, Pass\}, and in DeepSeek the same group appears via the smooth-onset \texttt{build\_and\_hold} shape; in Rust the only Qwen \texttt{two\_phase} concepts are Super and Use, both module-system keywords. Calibration negatives behave correctly: Python builtins have near-zero concept fractions (Qwen $0.024$, DeepSeek $0.068$), and Rust ownership operators (\texttt{\&}, \texttt{*}, \texttt{'}), where token and concept are inseparable, also return near-zero.

\subsection{The ``What'' Is Conserved}
\label{sec:what}

\begin{figure*}[!t]
    \centering
    \begin{subfigure}[t]{0.48\linewidth}
        \includegraphics{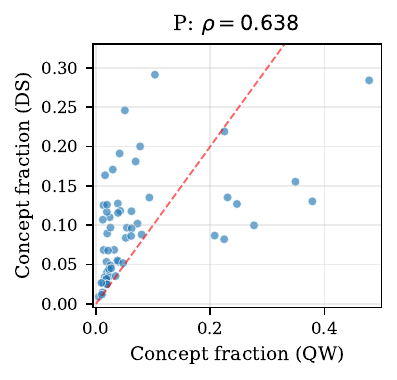}
        \caption{Python ($\rho = 0.638$, $n = 58$).}
    \end{subfigure}\hfill
    \begin{subfigure}[t]{0.48\linewidth}
        \includegraphics{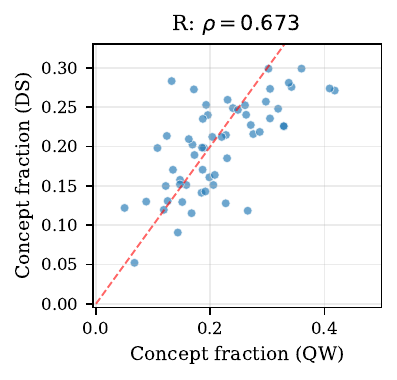}
        \caption{Rust ($\rho = 0.673$, $n = 57$).}
    \end{subfigure}
    \caption{Concept fraction Qwen vs DeepSeek, one point per concept, with $y = x$ reference. Rankings are conserved across models (\S\ref{sec:what}).}
    \label{fig:concept_scatter}
\end{figure*}

\textbf{Which constructs earn dedicated circuitry is task-determined and transfers across models.} Concept-fraction rankings agree (Figure~\ref{fig:concept_scatter}, Table~\ref{tab:correlations}): Spearman $\rho = 0.638$ for Python and $0.673$ for Rust (both $p < 10^{-7}$), with flow-type classifications agreeing at 88.6\% and 85.1\%. The ranking depends on the language and the prediction task, not on the particular model. The correlations are moderate, not near-unity --- the unexplained variance is the model-specific component the next two axes characterise.

\begin{table}[!t]
\centering
\footnotesize
\setlength{\tabcolsep}{4pt}
\begin{tabular}{@{}lrrlc@{}}
\toprule
Language & $n$ & $\rho$ & $p$ & Flow-type agr.\ \\
\midrule
Python & 58 & 0.638 & $< 10^{-7}$ & 88.6\% \\
Rust   & 57 & 0.673 & $< 10^{-7}$ & 85.1\% \\
\bottomrule
\end{tabular}
\caption{Cross-model concept-fraction ranking (Spearman $\rho$) and \S\ref{sec:arch:atomicity} flow-type assignment agreement, per language (\S\ref{sec:what}).}
\label{tab:correlations}
\end{table}

\subsection{The ``Where'' Diverges}
\label{sec:where}

\begin{figure*}[!t]
    \centering
    \includegraphics{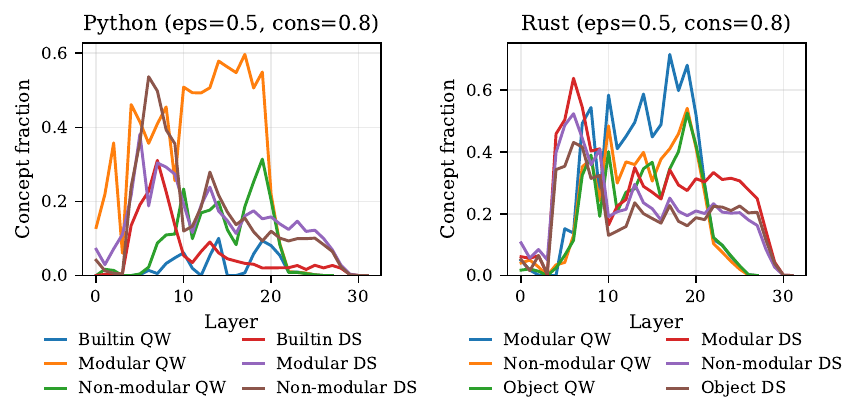}
    \caption{Layer-resolved concept-fraction profiles. Qwen peaks in a late band ($\sim$L17--19); DeepSeek peaks at L6--7 in both languages (\S\ref{sec:where}).}
    \label{fig:concept_fraction_profile}
\end{figure*}

\textbf{Where a construct is processed is a property of the model, not the language.} Qwen concentrates concept-specific processing in a late band (Python $\sim$L19, Rust $\sim$L17); DeepSeek at L6--7 for both languages (Figure~\ref{fig:concept_fraction_profile}), roughly 12--13 layers earlier. The within-model cross-language consistency is the load-bearing observation: each model's band predicts where it processes both languages. This concerns which layer carries the cleanest concept-specific signal, not where most representational activity lives --- that is circuit size, below.

\subsection{The ``How'' Diverges}
\label{sec:how}

\begin{figure*}[!t]
    \centering
    \includegraphics{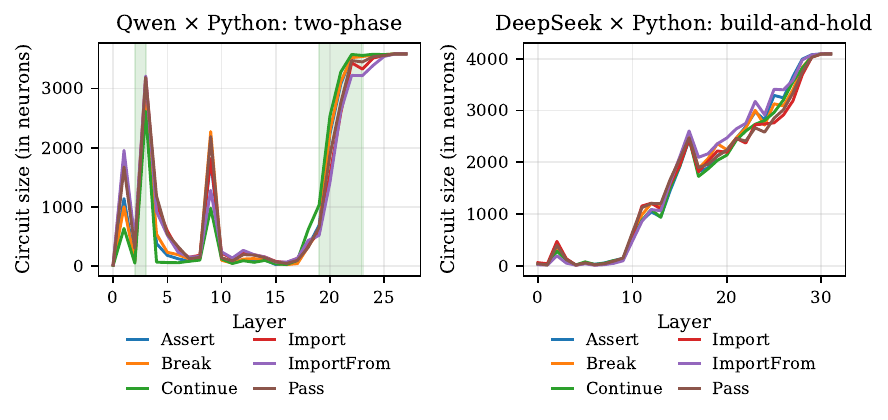}
    \caption{Circuit size by layer for the six Python atomicity concepts. Qwen shows an early spike at L2--3 absent in DeepSeek; DeepSeek climbs smoothly from L10 (\S\ref{sec:how}).}
    \label{fig:atomicity_dynamics}
\end{figure*}

\textbf{The same atomic concepts receive an early circuit spike in Qwen that they do not in DeepSeek.} This claim lives on circuit size, a different metric from the concept-fraction peaks of \S\ref{sec:where}. Figure~\ref{fig:atomicity_dynamics} plots circuit size by layer for the six Python atomicity concepts: Qwen shows a sharp early spike at layers 2--3 (early-bias $0.73$--$0.90$), while DeepSeek's circuits are near-empty until layers 10--12 then climb smoothly (early-bias $0.05$--$0.11$). The presence versus absence of the early spike is the divergence. In Qwen the late-layer mask approaches the full MLP dimension --- saturation, not a meaningful peak --- so the robust contrast is the early-layer difference, which the early-bias measure (normalising early against late) captures. Across three architectures the same concepts show three signatures: an inverted-U in the sparse CSP-Atlas model, an early spike in Qwen, smooth late onset in DeepSeek.

\subsection{Interpretation}
\label{sec:interpretation}

\textbf{Circuit structure has at least three independent axes, and ``are circuits universal?'' conflates them.} Which concepts earn circuitry is task-determined and transfers. Where they are processed, and how their circuits build over layers, are model-determined and do not. A circuit can be universal in concept identity while non-universal in processing depth and dynamics. The answer to the universality question is: yes on What, no on Where and How.

This refines the universality hypothesis \citep{Olah2020,Gurnee2024,Huh2024}, which holds that independently trained models converge to shared representations. We find the convergence is axis-dependent: it holds for \emph{what} earns circuitry --- representational content --- but not for \emph{where} or \emph{how} that circuitry is organised. Universality is a property of content, not of computational organisation.

Two statements bound the interpretation. First, the ``-determined'' terms are defined within this 2$\xtimes$2: task-determined means ``constant across the model axis of this design''; model-determined means ``constant across the language axis and varying across the model axis''. With two models, ``model-determined'' attributes the variation to whatever distinguishes the two training runs --- corpus, trajectory, architecture, initialisation --- without isolating which factor; isolating it requires designs that vary one factor at a time. Second, one may speculate that the divergence reflects a loss landscape with multiple equally good solutions that agree on representational content while differing in computational strategy --- but this is speculation: the study measures representations, not the loss landscape.

\section{Language Design Determines Representation Strength}
\label{sec:lang}

\subsection{Rust vs Python}
\label{sec:lang:strength}

\textbf{Rust constructs receive 2--3$\xtimes$ more concept-specific circuitry than Python equivalents, in both models and the same direction --- a property of language design, not architecture.} Mean concept fraction is Qwen $0.215$ vs $0.074$ ($2.91\xtimes$) and DeepSeek $0.201$ vs $0.097$ ($2.07\xtimes$) (Table~\ref{tab:cf_by_group}). In Python (Qwen) a sharp three-tier hierarchy separates modular keywords (atomic, no nested body: \texttt{import}, \texttt{assert}, \texttt{break}, \texttt{continue}, \texttt{pass}) at $0.314$, non-modular control-flow constructs at $0.090$, and builtins at $0.024$ --- a 13:1 ratio. These three tiers sit within the testable subset, which by construction excludes the tokenless-AST tier of Wilam (2026) --- tokenless constructs have no keyword and so no concept-versus-token contrast. In Rust the three groups (Modular, Non-modular, Object) fall within $1.5\xtimes$ of one another: stricter syntax makes all constructs distinctive, so the hierarchy compresses. The sharp modular-keyword tier is itself Qwen-specific: in Python under DeepSeek the order inverts, with non-modular constructs ($0.151$) edging out modular ones ($0.144$) (Table~\ref{tab:cf_by_group}). (Counts: Python $6 + 18 + 34 = 58$; Rust $6 + 15 + 36 = 57$.)

\begin{table}[!t]
\centering
\small
\begin{tabular}{@{}lrrrr@{}}
\toprule
Group & P$\xtimes$QW & P$\xtimes$DS & R$\xtimes$QW & R$\xtimes$DS \\
\midrule
Modular     & 0.314 & 0.144 & 0.294 & 0.264 \\
Non-modular & 0.090 & 0.151 & 0.222 & 0.213 \\
Builtin / Object & 0.024 & 0.060 & 0.200 & 0.185 \\
\bottomrule
\end{tabular}
\caption{Mean concept fraction by group, per (language, model) cell. The Python (Qwen) tier hierarchy (13:1 ratio) compresses in Rust (within $1.5\xtimes$) and inverts in Python (DeepSeek), where non-modular slightly exceeds modular (\S\ref{sec:lang:strength}).}
\label{tab:cf_by_group}
\end{table}

\subsection{The Boost Is Largest for Python's Weakest Concepts}
\label{sec:lang:boost}

\textbf{The boost --- the ratio of a concept's Rust to Python concept fraction --- is largest where Python is weakest.} Concepts already strong in Python (While, Break, Continue) show modest boosts; concepts weak in Python (Return $7.2\xtimes$, For $5.1\xtimes$, If $4.4\xtimes$ --- high token ambiguity) receive the largest. Import/Use is the one case where the ratio falls below 1: Python's \texttt{import} has more distinctive syntax than Rust's \texttt{use}. This follows the token-ambiguity principle of Wilam (2026): concept-only fraction is driven by how ambiguous a keyword's surface token is, and Rust's stricter syntax reduces that ambiguity for constructs like \texttt{for} and \texttt{if}, forcing structural representations. The fraction is predictable from token ambiguity paired with structural distinctiveness --- across languages, not just within one.

\subsection{Cross-Language Neuron Sharing}
\label{sec:lang:sharing}

\begin{figure}[!t]
    \centering
    \includegraphics{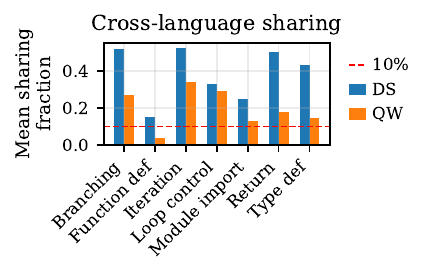}
    \caption{Cross-language neuron sharing by equivalence class. DeepSeek shares $1.94\xtimes$ more than Qwen; the $10\%$ threshold is shown as a dashed line (\S\ref{sec:lang:sharing}).}
    \label{fig:cross_language_sharing}
\end{figure}

\textbf{Both models share neurons for equivalent constructs across Python and Rust; what is shared is conserved, how much is model-determined.} We select seven cross-language pairings of constructs that play equivalent roles --- iteration, branching, loop control, function definition (\texttt{def}$\leftrightarrow$\texttt{fn}), module import, return, type definition (Table~\ref{tab:cross_lang_pairings}) --- and measure whether the model shares neurons between them. A pairing shares if its sharing fraction (intersection of pooled concept-only sets over the smaller pool, averaged over layers) exceeds 10\%; passing pairings sit well above this and the one miss well below, so the count is insensitive to the cutoff. DeepSeek shares $1.94\xtimes$ more cross-language than Qwen (Figure~\ref{fig:cross_language_sharing}). All seven pass in DeepSeek; Qwen reaches six, missing function definition at $3.8\%$ --- the model does not treat indentation-scoped \texttt{def} and brace-scoped \texttt{fn} as equivalent despite their shared role, the surface forms differing too much. The ranking of how shareable a pairing is, is conserved across models (task-determined); the magnitude difference is model-determined, plausibly reflecting the \S\ref{sec:how} dynamics, though the present data does not formally test this.

\begin{table}[!t]
\centering
\small
\begin{tabular}{@{}lrrcc@{}}
\toprule
Equivalence class & DS & QW & DS pass & QW pass \\
\midrule
Iteration       & 0.525 & 0.340 & \checkmark & \checkmark \\
Branching       & 0.517 & 0.271 & \checkmark & \checkmark \\
Return          & 0.503 & 0.177 & \checkmark & \checkmark \\
Type def        & 0.434 & 0.146 & \checkmark & \checkmark \\
Loop control    & 0.330 & 0.289 & \checkmark & \checkmark \\
Module import   & 0.250 & 0.129 & \checkmark & \checkmark \\
Function def    & 0.149 & 0.038 & \checkmark & --         \\
\bottomrule
\end{tabular}
\caption{Cross-language sharing fraction (per layer mean) for the seven equivalence classes. Threshold for "passes" is $\geq 10\%$. DeepSeek passes all seven; Qwen reaches six (function definition fails) (\S\ref{sec:lang:sharing}).}
\label{tab:cross_lang_pairings}
\end{table}

\section{Architecture-Invariant Organisation}
\label{sec:arch}

\subsection{The Atomicity Super-Cluster}
\label{sec:arch:atomicity}

\textbf{The atomicity super-cluster replicates across three architectures; the conserved set is task-determined, its temporal shape architecture-specific.} In Qwen the Python \texttt{two\_phase} concepts are exactly the six-concept CSP-Atlas atomicity set \{Assert, Break, Continue, Import, ImportFrom, Pass\}. In DeepSeek the group appears via \texttt{build\_and\_hold}; the Rust analogues are Super and Use. The replication spans sparse 8-layer, dense 28-layer, and dense 32-layer models, with a different temporal shape in each. Flow-type agreement between the dense models is $88.6\%$ (Python) and $85.1\%$ (Rust). Appendix Figure~\ref{fig:circuit_size} shows per-layer profiles for all four cells.

\subsection{Semantic Clustering in Rust}
\label{sec:clustering}

\textbf{Qwen represents Rust's type-trait family as a strongly bound neuron cluster, recovering a type-theoretic dimension invisible at the surface level --- but selectively.} Hierarchical clustering at Qwen $\xtimes$ Rust $\xtimes$ L14 (Figure~\ref{fig:cluster_dendrogram}) suggests four groups; a permutation test (10{,}000 same-size random draws, fixed seed; Figure~\ref{fig:cluster_perm_test}) sorts them:

\begin{figure*}[!t]
    \centering
    \includegraphics[width=\linewidth]{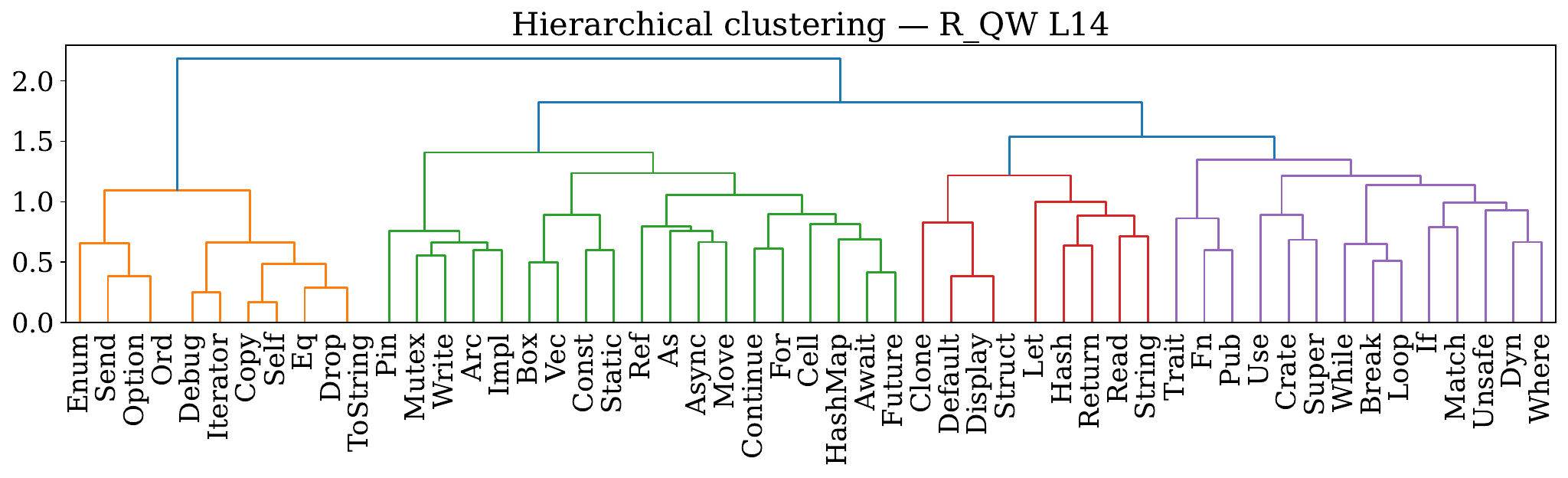}
    \caption{Rust semantic clustering at Qwen $\xtimes$ L14 --- Ward dendrogram on $1-\text{Jaccard}$ over the non-empty concept-only sets. The four hypothesised groups (type-system traits, memory/ownership, data definition, control-flow/module) are tested for cohesion in Figure~\ref{fig:cluster_perm_test} (\S\ref{sec:clustering}).}
    \label{fig:cluster_dendrogram}
\end{figure*}

\begin{figure}[!t]
    \centering
    \includegraphics{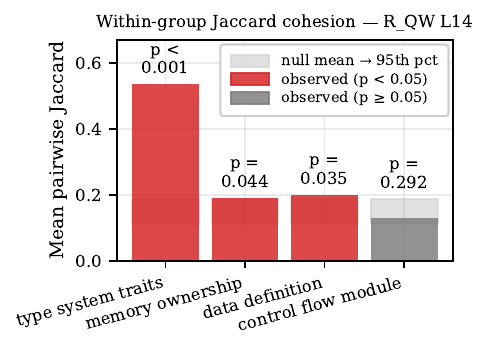}
    \caption{Permutation test (10{,}000 same-size random draws, seed = 42) on the four hypothesised groups from Figure~\ref{fig:cluster_dendrogram}: one strongly bound cluster (type-system traits, $p < 0.001$), two marginal, one random.}
    \label{fig:cluster_perm_test}
\end{figure}

\begin{table}[H]
\centering
\footnotesize
\setlength{\tabcolsep}{4pt}
\begin{tabular}{@{}lcccl@{}}
\toprule
Group & Jacc. & Null & $p$ & Verdict \\
\midrule
Type-system traits   & 0.535 & 0.112 & $<\!0.001$ & strong \\
Memory/ownership     & 0.190 & 0.112 & 0.044 & marg. \\
Data definition      & 0.199 & 0.112 & 0.035 & marg. \\
Control-flow/module  & 0.129 & 0.112 & 0.292 & random \\
\bottomrule
\end{tabular}
\end{table}

The four hypothesised groups: \emph{type-system traits} (Enum, Send, Option, Iterator, Copy, Eq, Drop, Debug, ToString), \emph{memory/ownership} (Pin, Mutex, Arc, Box, Vec, Impl, Async, Move, Future), \emph{data definition} (Struct, Let, String, Trait, Default, Display, Hash, Return, Read), \emph{control-flow/module} (Fn, Pub, Use, Crate, While, Break, Loop, If, Match). The honest finding is one strongly bound cluster, two marginally significant, one random --- not four equal clusters. The type-trait family is the striking case: nine keywords spanning four syntactic categories (marker traits, method-carrying traits, a sum-type keyword, a generic enum), which the model collapses on the basis of their shared role in Rust's type-and-trait machinery --- a dimension absent from surface syntax and the AST. Its cohesion is more than double the marginal groups. The control-flow negative result is informative: strong specialised machinery for the type system, none detectable for control flow, though both are syntactically distinctive. For Python at the same layer (Appendix Figure~\ref{fig:python_dendrogram_qwen}) control-flow keywords \emph{do} cluster, so the model's neuron-level organisation of control flow differs by language, sharpening rather than diluting the type-trait finding.

\subsection{Meta-Circuit Structure}
\label{sec:meta}

\textbf{Rust's richer circuitry produces more coordinated meta-circuit organisation, and DeepSeek's smoother dynamics more than Qwen's.} DeepSeek produces roughly $3\xtimes$ more statistically significant meta-circuit structure; Rust meta-circuits are broader than Python in both models (Rust Loop control, Iteration, Branching show 5--7 significant layers each versus Python's 0--7). This coheres with both the language-design effect and the processing-style difference.

\section{Validation}
\label{sec:validation}

\subsection{Causal Validation: Double Dissociation}
\label{sec:validation:ablation}

Per-layer zero-ablation of concept-only neurons on Qwen Python, on concept and matched checker prompts, against a size-matched random null, following the activation-patching tradition \citep{Meng2022,Conmy2023}. A clean double dissociation requires concept-only ablation to drop the target probability more on concept than checker prompts, and more than the random null.

\begin{figure}[!t]
    \centering
    \includegraphics{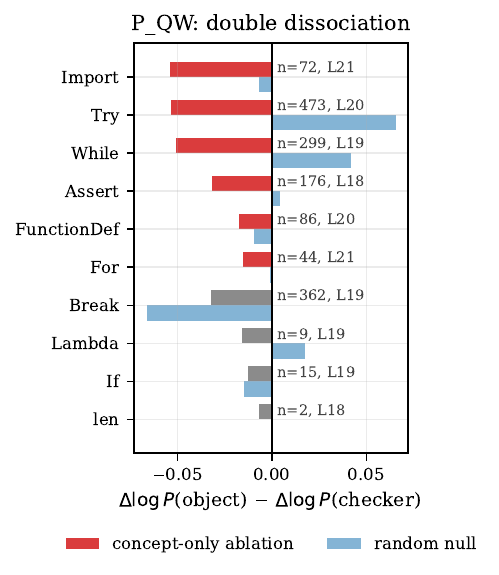}
    \caption{Double dissociation by concept at the peak-effect layer. Four concepts (Import, Try, While, Assert) pass with concept-only ablation more dissociative than random null; Break fails; small-$n$ concepts show no measurable effect (\S\ref{sec:validation:ablation}).}
    \label{fig:dissociation}
\end{figure}

Six concepts satisfy the direction of the dissociation. Four pass with clear margins: Import, Try, While, Assert (Figure~\ref{fig:dissociation}, Table~\ref{tab:dissociation}), with peak-effect layers inside the concept-fraction band (L18--21) --- the Where signature, replicated causally --- and spanning two clusters identified correlationally: the atomicity super-cluster (Import, Assert) and body-scoped control flow (Try, While). Two more --- FunctionDef ($-0.018$ against a null of $-0.009$) and For ($-0.015$ against $-0.001$) --- satisfy the direction with thin margins, and we mark them marginal rather than counting them as clean passes.

One concept fails outright: Break, whose concept-only ablation ($-0.032$) produces a smaller dissociation than the random null ($-0.066$) despite a large concept-only set; \texttt{break} is token-rare, drawing on general lexical context rather than a concept-selective pathway, so ablating any large set at L19 hurts it similarly. Concepts with concept-only sets below $\sim$20 neurons show no measurable effect, consistent with the probe result below.

\noindent\textbf{Scope of the causal claim.} This validation covers one of the four cells: Qwen $\xtimes$ Python, deliberately the cell with the clearest two-phase profile, where single-layer ablation is best defined. DeepSeek's build-and-hold profile has no clear peak layer, so validating it requires an extended design --- multi-layer or cumulative ablation, not a rerun. Until then, the causal evidence supports the decomposition where tested, and the cross-model claims rest on the correlational and geometric evidence.

\begin{table}[!t]
\centering
\footnotesize
\setlength{\tabcolsep}{3pt}
\begin{tabular}{@{}lrrrrc@{}}
\toprule
Concept & $n$ & L & $\Delta_{\text{co}}$ & $\Delta_{\text{null}}$ & P \\
\midrule
Import       & 72  & 21 & $-0.054$ & $-0.007$ & \checkmark \\
Try          & 473 & 20 & $-0.053$ & $+0.066$ & \checkmark \\
While        & 299 & 19 & $-0.051$ & $+0.042$ & \checkmark \\
Assert       & 176 & 18 & $-0.032$ & $+0.004$ & \checkmark \\
FunctionDef  & 86  & 20 & $-0.018$ & $-0.009$ & $\sim$ \\
For          & 44  & 21 & $-0.015$ & $-0.001$ & $\sim$ \\
Break        & 362 & 19 & $-0.032$ & $-0.066$ & $\times$ \\
\bottomrule
\end{tabular}
\caption{Per-concept double dissociation, Qwen $\xtimes$ Python. $\Delta$ is the change in $\log P(\text{target})$ between concept and matched checker prompts at the peak-effect layer; concept-only ablation must drop the target more than a size-matched random null. Four pass with clear margins; FunctionDef and For are marginal ($\sim$); Break fails despite a large concept-only set (\S\ref{sec:validation:ablation}).}
\label{tab:dissociation}
\end{table}

\subsection{Geometric Validation: Linear Probes}
\label{sec:validation:probes}

Per-concept, per-layer logistic-regression probes classifying concept from checker prompts achieve $97.6$--$99.7\%$ accuracy at every one of Qwen's 28 layers, for the 24 keyword-bearing AST concepts (Figure~\ref{fig:probe_validation}). The distinction is linearly decodable from the earliest layers, independent of any threshold. Pairwise probe-direction cosine similarity correlates with pairwise concept-only Jaccard similarity across 276 concept pairs, Pearson $r$ peaking at $0.645$ at L20. The correlation is moderate, not unity: the two views agree on coarse structure but disagree on low-amplitude tails, the probe direction spanning many neurons below $\eps = 0.5$ that the binary view does not capture. Where concept-only sets are large, ablation produces clean dissociations; where small, ablation has no effect but probes still exceed $97\%$ --- the representation exists and is decodable but lives in a sub-threshold regime the binary method does not reach. The decomposition is a correct but partial view, and the moderate correlation states where the two regimes meet.

\section{Discussion}
\label{sec:discussion}

\paragraph{Two processing styles as a model fingerprint.} The findings yield a per-model profile (Table~\ref{tab:fingerprint}): six independent measurements --- concept-fraction peak band, atomicity flow type, Rust cluster strength, meta-circuit significance rate, mean cross-language sharing, and the cross-model flow-type agreement --- all separate or track the two models the same way. Qwen's late, sharp concentration comes with fine-grained clustering and lower cross-language overlap; DeepSeek's early, smooth growth comes with coarser clustering and more overlap. Six measurements could have varied independently; they cohere into two styles. That is the dimensionality finding: cross-model variation, at least here, is low-dimensional.

\paragraph{Scope of the fingerprint, and the predictions it licenses.} Two models cannot show that the fingerprint \emph{predicts}: any pair of independently trained models will differ somehow, and coherence observed once could in principle be coincidence. What the fingerprint licenses is a set of falsifiable predictions for any third model --- its peak band, atomicity flow type, cluster granularity, and cross-language sharing level should co-vary as one style rather than mixing freely. A model with an early peak band, for example, should show smoother atomicity onsets and higher cross-language sharing. Testing this on a third model is the designed next step; the present contribution is that the 2$\xtimes$2 defines the fingerprint and shows its internal coherence, which is what makes the prediction well-posed.

\begin{table}[!t]
\centering
\footnotesize
\setlength{\tabcolsep}{3pt}
\begin{tabular}{@{}p{0.34\columnwidth}p{0.27\columnwidth}p{0.27\columnwidth}@{}}
\toprule
Measurement & Qwen & DeepSeek \\
\midrule
Conc.-fraction peak    & late (L19/L17)        & early (L6--7) \\
Atomicity flow (P)     & two-phase             & build-and-hold \\
Rust type-trait (L14)  & $J{=}0.535$ ($p\!<\!0.001$) & not detected \\
Meta-circuit signif.   & baseline              & $\sim$3$\xtimes$ more \\
Mean x-lang sharing    & 0.199                 & 0.387 \\
\midrule
Flow-type agr.\ (P/R)  & \multicolumn{2}{c}{88.6\% / 85.1\% (QW$\leftrightarrow$DS)} \\
\bottomrule
\end{tabular}
\caption{Per-model fingerprint: five per-model contrasts separating Qwen vs.\ DeepSeek the same way --- late/sharp vs.\ early/smooth processing style --- plus the cross-model flow-type agreement (bottom, a pairwise quantity) (\S\ref{sec:discussion}).}
\label{tab:fingerprint}
\end{table}

\paragraph{The language axis is orthogonal to the model axis.} The language-design effect --- Rust's higher concept-specificity and the cross-language sharing pattern --- holds in both models and the same direction, while the models differ on where and how they process concepts. The $2\xtimes2$ design was built to detect this separability, and does: what earns circuitry is task-determined, where and how it is processed is model-determined, how strongly a construct is represented is language-determined --- three independent sources of variation.

\paragraph{What this enables, operationally.} Concept-level knowledge transfers: the $\rho \approx 0.65$ rankings and the conserved sharing structure mean an inventory measured on one model is a usable prior for another. Layer-level knowledge does not: the 12--13 layer offset between the models' bands means a fixed-depth probe or patch tuned on one model targets the wrong layers on the other. A practitioner porting an interpretability result should ask which axis it lives on; the fingerprint then says what must be re-measured (the band, the flow type) and what can be reused (the inventory, the pairings).

\paragraph{The method is generative.} Measuring a structured inventory exhaustively surfaces more regularities than one paper can pursue. We develop the three bearing on the master question, but the released artifacts contain further structure --- the asymmetry in meta-circuit organisation, the per-language difference in control-flow grouping, the layer-band structure of concept-specific processing. The wide first stage surfaces; deeper stages and other work can mine.

\paragraph{Future work.} A continuous-geometry extension, in progress, would capture the sub-threshold signal the binary view misses. The natural target is natural language --- the method's eventual home, where the inventory is less clean but the questions matter most. A developmental analysis over training checkpoints would ask when the What/Where/How split appears and whether What is conserved while Where and How drift. And formal languages with strict internal logic --- a proof assistant such as Lean, where constructs carry logical content --- would test whether the method reaches a qualitatively different kind of abstraction.

\section*{Limitations}

\noindent\textbf{Sub-threshold structure.} The method captures high-amplitude neurons and misses distributed sub-threshold structure, quantified in \S\ref{sec:validation:probes}; the continuous treatment is deferred.

\noindent\textbf{Imperative languages only.} Both languages are imperative; generalisation to declarative languages or proof assistants is open.

\noindent\textbf{Layer-count mismatch.} Layer counts differ (28 vs 32); body comparisons use absolute indices, appendix profiles fraction-of-depth.

\noindent\textbf{Causal validation scope.} Causal validation is on Qwen Python only.

\noindent\textbf{Degenerate consistency parameter.} The consistency parameter, which varies meaningfully in the sparse model of Wilam (2026), is degenerate for these dense SwiGLU architectures, so the effective sweep is over $\eps$ alone.

\noindent\textbf{Cross-model concept space.} The full concept space differs across models; comparisons use the shared testable subsets.

\section*{Ethical Considerations}

This work analyses internal representations of publicly available open-source models. No new models are trained, no human subjects are involved, no private data is used. Prompts are synthetic code generated programmatically.

\paragraph{Licenses and intended use.} Qwen2.5-Coder-7B is distributed under the Apache-2.0 license; DeepSeek-Coder-V1-6.7B under the DeepSeek License Agreement. Both licenses permit research and commercial use, and our research-only use of these models is consistent with their stated terms. The code released with this paper uses Apache-2.0, and the dataset uses CC-BY-4.0; both are intended for research and commercial re-use with attribution, compatible with the upstream model licenses.

\section*{Availability}

Analysis code, figure scripts, and the frozen-numbers test suite are released at \url{https://github.com/piotrwilam/Atlas2x2} (release \texttt{v1.0.0}, Apache-2.0). The frozen experimental artifacts are released as a dataset at \url{https://huggingface.co/datasets/piotrwilam/Atlas2x2} (CC-BY-4.0); the loaders in \texttt{atlas/io/} auto-fetch missing files on demand. Every number and figure in the paper regenerates from the released analysis layer without rerunning a model; see Appendix~\ref{app:reproducibility}.


\clearpage

\onecolumn
\appendix

\section{Reproducibility and Claim Verification}
\label{app:reproducibility}

Code lives at \url{https://github.com/piotrwilam/Atlas2x2} (release \texttt{v1.0.0}); data lives at \url{https://huggingface.co/datasets/piotrwilam/Atlas2x2}, and missing files auto-fetch on demand. Every numerical claim in the paper is regenerable from this release. Running \mbox{\texttt{pytest tests/}}\allowbreak\mbox{\texttt{test\_paper\_numbers.py}} verifies all locked numbers. Each figure regenerates from a config-named script under \texttt{experiments/}, e.g.\ \mbox{\texttt{python experiments/}}\allowbreak\mbox{\texttt{fig2\_concept\_scatter.py}} \mbox{\texttt{--config-name paper/figure2\_concept\_scatter}} reproduces Figure~\ref{fig:concept_scatter}.

\paragraph{Compute and infrastructure.} The model-touching stages (extraction, $\eps$-sweep, probes, ablation) run on a single NVIDIA A100 GPU (40GB) via Google Colab Pro+. Extraction is the dominant cost: $\sim$160k forward passes per cell across 115 concepts (58 Py + 57 Ru) $\times$ 50 prompts $\times$ 2 sets, repeated for 4 cells (Python/Rust $\times$ Qwen/DeepSeek), totalling $\sim$24 GPU-hours. Linear probes (\S\ref{sec:validation:probes}) and zero-ablation (\S\ref{sec:validation:ablation}) add $\sim$2 GPU-hours combined. All analysis, plotting, and the locked paper-number tests run on CPU in seconds.

\newcommand{\testname}[1]{\texttt{\seqsplit{#1}}}

\begin{table}[H]
\centering
\scriptsize
\setlength{\tabcolsep}{3pt}
\renewcommand{\arraystretch}{1.15}
\begin{tabular}{@{}p{0.12\columnwidth}p{0.28\columnwidth}p{0.50\columnwidth}@{}}
\toprule
Section & Claim & Value \,/\, Verifies via \\
\midrule
\S\ref{sec:method} & Testable concept counts & Py 58 ($6{+}18{+}34$), Ru 57 ($6{+}15{+}36$) \newline \testname{f2\_concept\_group\_counts} \\
\S\ref{sec:what} & Cross-model $\rho$ & 0.638 (Py), 0.673 (Ru); $p<10^{-7}$ \newline \testname{f1\_concept\_fraction\_spearman} \\
\S\ref{sec:how}/\S\ref{sec:arch:atomicity} & Atomicity flow types & 6 concepts \texttt{two\_phase}; agr.\ 88.6\%/85.1\% \newline \testname{f6\_1\_flow\_type\_agreement\_between\_models}, \testname{p\_qw\_two\_phase\_is\_exactly\_the\_atomicity\_super\_cluster} \\
\S\ref{sec:lang:strength} & Rust/Python strength ratio & 2.91$\xtimes$ (QW), 2.07$\xtimes$ (DS) \newline \testname{f3\_rust\_over\_python\_concept\_fraction\_ratios} \\
\S\ref{sec:lang:sharing} & Cross-language sharing ratio & 1.94$\xtimes$ (DS/QW); 7/7 DS, 6/7 QW \newline \testname{f3\_cross\_language\_sharing\_ratio} \\
\S\ref{sec:clustering} & Type-trait cluster cohesion & Jaccard 0.535, null 0.112, $p\!<\!0.001$ \newline \testname{f6\_g1\_trait\_family\_observed\_jaccard}, \testname{f6\_g1\_permutation\_p\_value} \\
\S\ref{sec:clustering} & Other cluster verdicts & G2 0.044, G3 0.035, G4 0.292 \newline \testname{f6\_four\_cluster\_observed\_and\_p\_value} \\
\S\ref{sec:validation:ablation} & Double dissociation & 4 pass; Break fails \newline \testname{f7\_1\_dissociation\_results} \\
\S\ref{sec:validation:probes} & Probe accuracy band & 97.6--99.7\%, 24 AST concepts \newline \testname{f7\_probe\_accuracy\_band}, \testname{f7\_probe\_accuracy\_exact\_endpoints} \\
\S\ref{sec:validation:probes} & Jaccard--cosine peak & $r\!=\!0.645$ at L20, 276 pairs \newline \testname{f8\_peak\_jaccard\_cosine\_correlation} \\
\bottomrule
\end{tabular}
\caption{Every paper-cited number is locked in \texttt{tests/test\_paper\_numbers.py} with $\pm 0.005$ tolerance; the typewriter line under each value gives the verifying test function name (prefix \texttt{test\_} omitted). Analysis routines --- correlations, Jaccard, permutation tests, flow-type classification --- live in \texttt{atlas/analysis/}; the model-touching pipeline (extraction, probes, ablation) lives in \texttt{circuits/}, separated because it requires a GPU.}
\label{tab:appendix_r}
\end{table}

\clearpage
\section{Per-Layer Circuit-Size Profiles}
\label{app:circuit_size}

\begin{figure}[H]
    \centering
    \begin{subfigure}[t]{0.48\linewidth}
        \includegraphics[width=\linewidth]{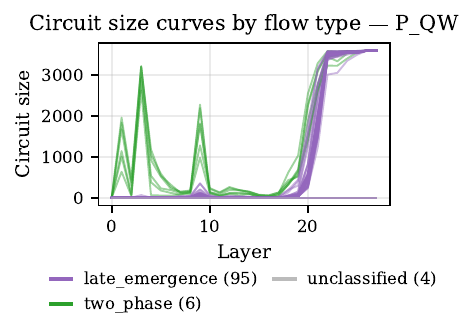}
        \caption{Python $\xtimes$ Qwen.}
    \end{subfigure}\hfill
    \begin{subfigure}[t]{0.48\linewidth}
        \includegraphics[width=\linewidth]{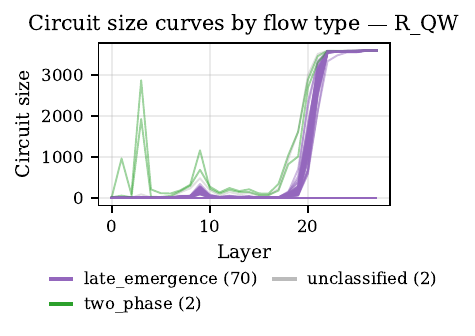}
        \caption{Rust $\xtimes$ Qwen.}
    \end{subfigure}\\[1ex]
    \begin{subfigure}[t]{0.48\linewidth}
        \includegraphics[width=\linewidth]{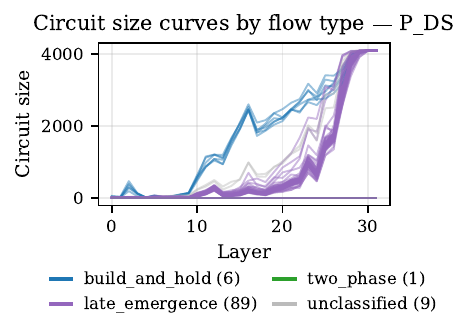}
        \caption{Python $\xtimes$ DeepSeek.}
    \end{subfigure}\hfill
    \begin{subfigure}[t]{0.48\linewidth}
        \includegraphics[width=\linewidth]{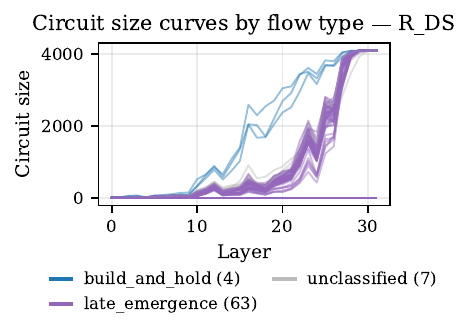}
        \caption{Rust $\xtimes$ DeepSeek.}
    \end{subfigure}
    \caption{Per-layer circuit size for every testable concept, coloured by flow type (\S\ref{sec:arch:atomicity}). Each panel shows one (language, model) cell.}
    \label{fig:circuit_size}
\end{figure}

Figure~\ref{fig:circuit_size} shows per-layer circuit size for every testable concept across all four cells, the full data behind the flow-type classifications summarised in \S\ref{sec:arch:atomicity}.

\clearpage
\section{Concept Clustering Across Models}
\label{app:concept_clustering}

\begin{figure}[H]
    \centering
    \includegraphics[width=\linewidth]{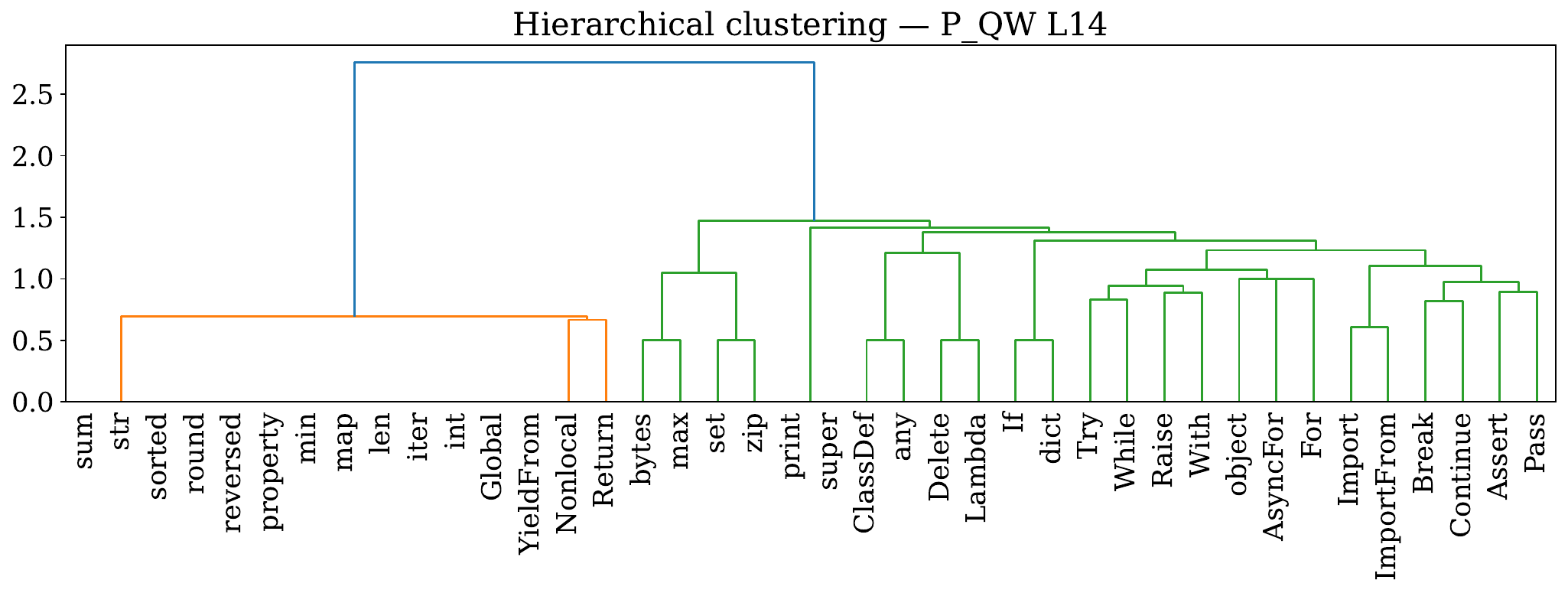}
    \caption{Python concept clustering, Qwen $\xtimes$ L14. The atomicity super-cluster and control-flow groupings emerge in the green tree (\S\ref{sec:clustering}).}
    \label{fig:python_dendrogram_qwen}
\end{figure}

\begin{figure}[H]
    \centering
    \includegraphics[width=\linewidth]{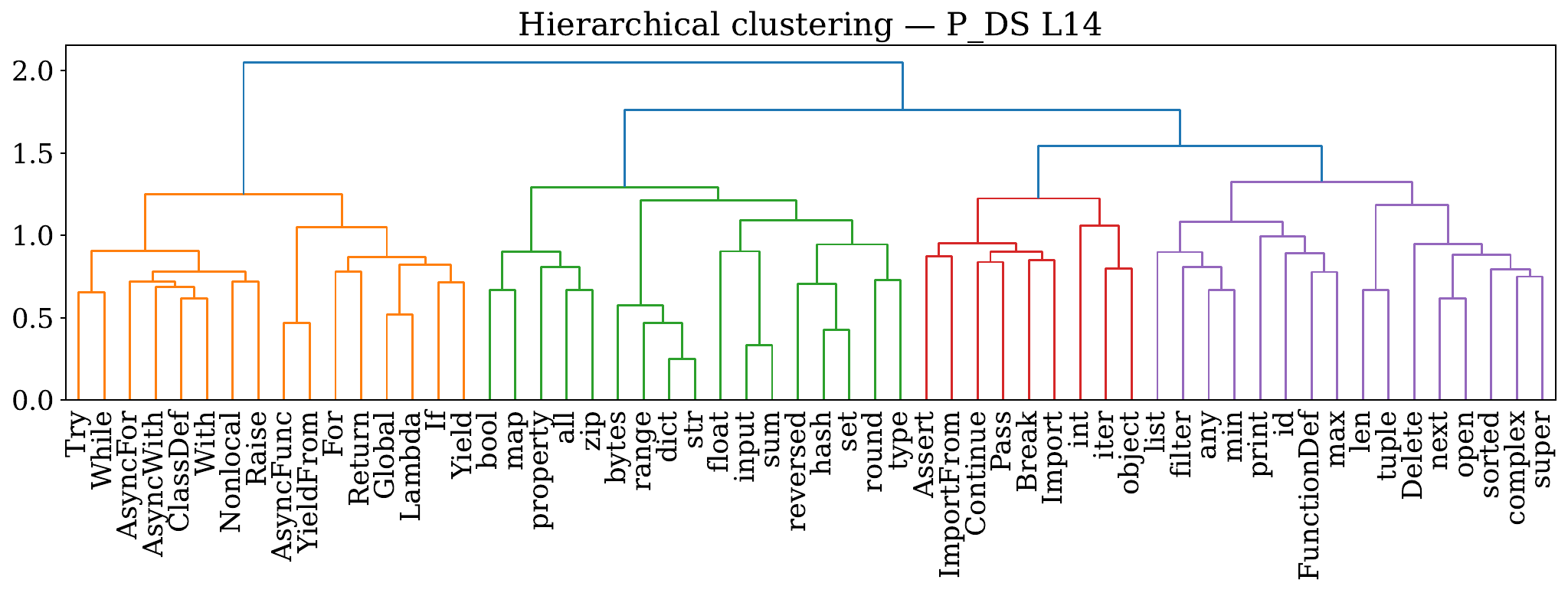}
    \caption{Python concept clustering, DeepSeek $\xtimes$ L14. Direct companion to Figure~\ref{fig:python_dendrogram_qwen} (same layer in both models for apples-to-apples comparison). The atomicity grouping replicates as a coherent cluster (\S\ref{sec:clustering}).}
    \label{fig:python_dendrogram_deepseek}
\end{figure}

Figures~\ref{fig:python_dendrogram_qwen} and~\ref{fig:python_dendrogram_deepseek} give the Python concept clustering for both models at L14, the cross-model companion to the Rust clustering in \S\ref{sec:clustering}; the atomicity grouping replicates in both.

\clearpage
\section{Linear-Probe Validation Detail}
\label{app:probe_validation}

\begin{figure}[H]
    \centering
    \begin{subfigure}[t]{0.34\linewidth}
        \includegraphics[width=\linewidth]{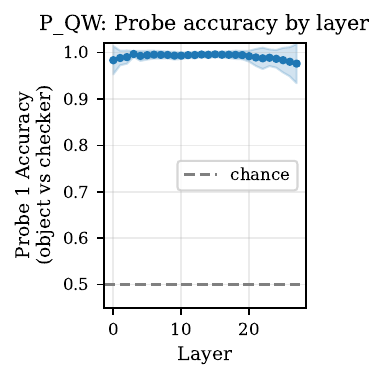}
        \caption{Probe accuracy by layer (97.6--99.7\%, 28 layers).}
    \end{subfigure}\hfill
    \begin{subfigure}[t]{0.63\linewidth}
        \includegraphics[width=\linewidth]{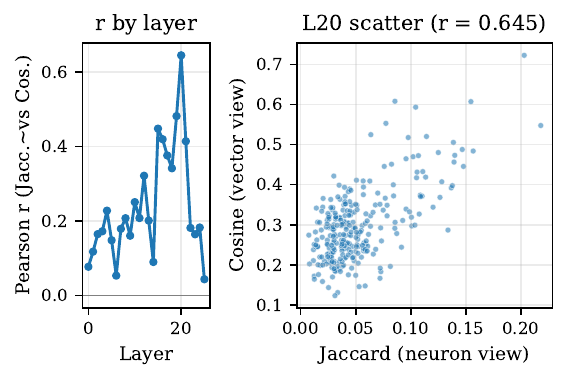}
        \caption{Jaccard--cosine correlation. Peak $r = 0.645$ at L20.}
    \end{subfigure}
    \caption{Linear-probe validation, Python $\xtimes$ Qwen, 24 AST concepts (\S\ref{sec:validation:probes}). Left: probe accuracy at every layer. Right: per-layer Pearson correlation between pairwise concept-only Jaccard and probe-direction cosine across 276 concept pairs, plus the L20 scatter where $r$ peaks.}
    \label{fig:probe_validation}
\end{figure}

Figure~\ref{fig:probe_validation} gives the per-layer probe accuracy and the Jaccard--cosine correlation underlying \S\ref{sec:validation:probes}.

\end{document}